# Evaluation protocols for historical script classification: the cost of manuscript-level leakage and the role of soft labels

D. I. Naryzhnyi, R. A. Popkov, D. A. Bakalin
ITMO University, St. Petersburg, Russia

## ABSTRACT

A paleographer applies a script type classifier to unseen manuscripts. We measure how much testing on other pages of training manuscripts inflates accuracy. On VML-HP-ext (715 pages, 171 Hebrew manuscripts, 14 script classes) we compare page-level and manuscript-level splits. Page-level accuracy drops from 97% to 54% (59% at manuscript level), and is 39% on the official blind test set. The gap of 40 to 47 percentage points (pp) at patch, page and manuscript level is statistically significant and reproduces with ResNet18; on binarized patches (ink and background only) it is 34 to 42 pp, also significant. On unseen manuscripts the model is overconfident: mean confidence 70% against 54% correct pages; patch-level expected calibration error (ECE, mean gap between confidence and accuracy) rises from 0.065 to 0.444. Binarization leaves patch-level overconfidence but cuts it markedly at page level. Confidence still helps triage: in the more confident half of pages 67% are correct, 41% in the rest. Training on expert soft labels showed no evidence of an accuracy difference, though a small page-level loss cannot be ruled out. The predicted distributions are closer to the expert's assessment. ECE is lower at patch level (0.21 to 0.31) and in most runs at page level, but not significantly at manuscript level. Label smoothing (blurring the label, no expert input) also lowers ECE but comes no closer to the expert. Temperature scaling (post-training confidence correction) helps significantly only at patch level. On 97 pages of 16 manuscripts with mixed expert assessment, soft labels cut divergence from the expert by 22 to 30% (significant in four of five runs); the point estimate of accuracy is 16.5 to 23.7 pp below hard labels, but with 16 manuscripts the loss can neither be confirmed nor ruled out. We propose recommendations for the evaluation protocol; code will be released with the paper.

## 1. INTRODUCTION

Digital paleography applies computer vision to tasks an expert traditionally does: script type classification, scribe attribution and, less often, dating. Open labeled datasets have made one formulation standard: the script type is predicted from a page image, and convolutional neural networks are the main tool.

A paleographer applies such a model to a manuscript that was not in training: previously unseen, poorly described or disputed. The paleographer needs to know how often the model is wrong on manuscripts of exactly that kind, and how far its confidence can be trusted. Many published figures answer a different question, because models are often tested on other pages of the very manuscripts they were trained on. A model can also recognize a manuscript it has already seen from the background pattern of its pages or from defects of the writing support. On our dataset, under that kind of evaluation, the model assigns the correct script type to 97% of pages, and to 54% of pages from manuscripts it has not seen. On unseen manuscripts it is also overconfident: mean confidence 70% against 54% correct. The agreement between confidence (the probability the model assigns to its predicted class) and accuracy is called calibration. The mismatch is measured by the expected calibration error (ECE): predictions are divided into groups of similar confidence, and the gap between confidence and accuracy is averaged over the groups. Finally, the expert assigns some manuscripts to several script types at once. If training gives every page a single class, that information is discarded. A label naming one class is called hard, whereas a soft label spreads weight over several classes in proportions taken from the expert's assessment (Section 3).

Published results in this area are hard to compare. Datasets and metrics vary across studies, and the splitting protocol adds a substantial discrepancy of its own. In several studies the training and test sets are not separated by manuscript (Section 2), so parts of one manuscript end up in both. The authors of VML-HP call random splitting the standard scheme and note that it can give misleading results [Droby et al., 2021]. A model trained that way can rely on manuscript-specific features and report accuracy that does not carry over to unseen manuscripts. When the test set holds pages of manuscripts the model already knows from training, this is called data leakage. We did not find a study that measures this effect on a public dataset of historical manuscripts, with identically built protocols and at several aggregation levels. The levels here are patches (individual parts of a page), pages and manuscripts.

We start from the VML-HP-ext dataset [Droby et al., 2022a; Droby et al., 2022b]. Its authors released the data with two test sets: the typical test set includes unseen pages of the manuscripts from the training set, while the blind test set consists of manuscripts that do not appear in the training set. On the previous version of the dataset, VML-HP, a sharp drop in accuracy was observed between typical and blind sets of the same kind [Droby et al., 2021; Vasyutinsky Shapira et al., 2022]. That drop, however, was measured on a single fixed split, and no confidence interval for the difference was given. For VML-HP-ext the authors of the dataset report results on the blind set only [Droby et al., 2022a; Droby et al., 2022b]. We generalize this observation: a grouped split by manuscript combined with cross-validation gives a distribution of estimates and confidence intervals, and it lets us measure the effect at three aggregation levels and check how far it depends on the choice of architecture.

Our contributions are as follows.

1. We quantify the "cost of leakage", that is, the difference in accuracy between a page-level and a manuscript-level split. The two protocols are built the same way, with five folds and each page in the test set exactly once. They differ in the variable used to form the folds. The estimate comes with bootstrap intervals and is obtained at three aggregation levels, and the effect is 40 to 47 percentage points of accuracy. With binarized patches (parts of the page mapped to two values, ink and background) it is significant as well and amounts to 34 to 42 percentage points.

2. We show that calibration degrades on unseen manuscripts, with patch-level ECE rising from 0.065 to 0.444, and that training on soft labels partly restores it. The gain is significant at the patch level in every run we checked. At the page level it is significant in three of the four soft-target runs against the original hard-label model (in eleven of the twelve pairs against its three runs). At the manuscript level the intervals for the differences include zero. Scalar temperature scaling calibrates better than soft labels at the patch level, but it gives no significant improvement at the page or manuscript level.
3. We compare training on hard labels with training on expert soft labels (cross-entropy with soft targets, KL divergence, mean squared error) on accuracy, closeness to the expert profile and calibration. We assess the differences with paired confidence intervals. We look separately at transitional pages, at a label-smoothing ablation and at the variation across training runs. For transitional pages we base the inference on the number of manuscripts, because the expert profile is defined per manuscript.
4. We also checked closeness to the expert's assessment without mapping the profile onto the fourteen classes. For this we trained a separate variant directly on the expert scores for regional styles and graphical modes. For regional styles we found no evidence of a difference between this variant and either soft cross-entropy or hard-label training, while for graphical modes it is closer to the expert's assessment than every other variant. This series does not bear on the conclusions about accuracy and calibration.
5. We will release the experiment code (the splitting protocols, training, and the computation of metrics and intervals) with the paper.

## 2. RELATED WORK

Among the classical approaches that rely on hand-crafted features, the Hinge and Fraglets descriptors and their derivatives are widely used. They have been applied to separating the scribes of the Great Isaiah Scroll from Qumran (1QIsaa) [Popović et al., 2021] and to dating the Dead Sea Scrolls [Popović et al., 2025]. The analysis in those studies runs over the columns of a single scroll and over images of 24 dated manuscripts. Convolutional models in Hebrew paleography were developed by the Visual Media Lab at Ben-Gurion University of the Negev: script type classification on the VML-HP (fifteen classes) [Droby et al., 2021; Vasyutinsky Shapira et al., 2022] and VML-HP-ext (fourteen classes) [Droby et al., 2022a] datasets, hard and soft labeling [Droby et al., 2022b], and multi-task learning for script classification and date estimation [Atamni et al., 2025]. The notion of typical and blind test sets was introduced in [Droby et al., 2021]. The VML-HP-ext dataset, with its official split and its expert soft labels, is described in [Droby et al., 2022a; Droby et al., 2022b]. We build on their data and their problem setting. Writer identification in historical documents was also studied in the Historical-WI competition [Fiel et al., 2017].

Traditional paleography also notes that the boundaries of hard script typologies are partly conventional: drawing on Hebrew manuscripts from the St. Petersburg collections, [Belkina, 2024] shows transitional zones between regional traditions, the transfer of script features by migrant scribes, and typologies that disagree across collection inventories. Seen from paleography itself, these observations argue for soft labeling.

Data leakage as a general problem of data analysis was formalized in [Kaufman et al., 2012], and in the taxonomy of [Kapoor, Narayanan, 2023] non-independence between train and test samples is singled out as a separate type of leakage. In clinical machine learning on wearable-sensor and smartphone data, record-wise splits often overestimate accuracy relative to subject-wise splits [Saeb et al., 2017].

In several studies of historical handwriting, the test images are taken from the same manuscripts as the training images, or the split does not take the manuscript into account: random selection of images in the Arabic KERTAS dataset [Adam et al., 2018], the dating of the Dead Sea Scrolls [Popović et al., 2025], and pages of a single manuscript, the Avila Bible [Cilia et al., 2020]. In writer retrieval one image is held out while the remaining images of the same document stay in the retrieval set [Christlein et al., 2022]. In the computational dating of Greek papyri, fragments of the same papyrus ended up in both the training and the test set [De Gregorio et al., 2024]. Unseen manuscripts in the test set are provided by the blind test sets of VML-HP and VML-HP-ext [Droby et al., 2021; Droby et al., 2022a] and by Test B in [Weißmann et al., 2024]. In the Historical-WI competition the writers of the test set do not appear in the training set [Fiel et al., 2017].

Comparisons of protocols on a single dataset have been published before. On VML-HP, seven convolutional networks correctly classified from 93.2% to 100% of the pages in the typical test set and from 26.7% to 39.7% of the pages in the blind test set. The authors attribute the gap to manuscript-specific features such as background texture [Droby et al., 2021; Vasyutinsky Shapira et al., 2022]. The authors of [Weißmann et al., 2024] likewise report that the networks learn codex-specific features. For Arabic manuscripts, the best configurations identified the writer correctly in 99.05% of cases when the data were split at the line level and in 78.61% when they were split at the page level. The two settings use different sets of writers (179 and 71) [Abushahla et al., 2026]. In these studies each protocol rests on a single split with its own test set, and no interval for the difference is reported. In our work the protocols are built the same way (five folds, each page in the test set exactly once) and differ in the grouping variable. We estimated the difference with a cluster bootstrap over manuscripts at three aggregation levels, and we measured calibration under both protocols.

Modern neural networks are often poorly calibrated, and on most of the datasets studied, temperature scaling restores calibration [Guo et al., 2017]. Label smoothing improves calibration as well, but it erases the information in the logits about resemblances between instances of different classes [Müller et al., 2019]. Training on distributions of human responses improves generalization to data that lie progressively further from the training distribution [Peterson et al., 2019]. The authors of the dataset used the VML-HP-ext expert soft labels for regression: a ResNet50 was trained with a mean squared error loss on the eight-dimensional vector of expert scores for regional styles and graphical modes. Once its output was converted into a class, accuracy on the blind test set was lower than with hard-label training [Droby et al., 2022b]. In [Droby et al., 2022a] the soft mode scores are used differently: they are turned into a squareness value for the script, and the graphical mode classes are redefined by that value. The expert profile lets us compare three ways of improving calibration on unseen manuscripts: temperature scaling, label smoothing and soft labels.

## 3. DATA

We used the VML-HP-ext v2 dataset [Droby et al., 2022a; Droby et al., 2022b; doi:10.5281/zenodo.6856934, CC-BY-4.0 license], which contains 715 page images excerpted from 171 medieval Hebrew manuscripts. The fourteen script classes are formed by combining six regional styles (Ashkenazi, Italian, Sephardic, Oriental, Byzantine, Yemenite) with graphical modes (square, semi-square and cursive). By graphical mode we mean here the way the script is executed within one regional style, running from square through semi-square to cursive. Four of the eighteen possible combinations are absent from the dataset: cursive for the Oriental and Byzantine styles, and semi-square and cursive for Yemenite. The official split is as follows: the training set holds 400 pages from 130 manuscripts, the typical test set 143 pages of the same 130 manuscripts, and the blind test set 172 pages from 41 manuscripts that do not appear in training. The training and typical test sets are therefore disjoint on the page level but share the same set of manuscripts, while neither of them shares a manuscript with the blind test set. The median number of pages per manuscript is four and the maximum is ten. Each page carries a hard class label and an expert soft label vector, which we call the expert profile. Each element of the vector is the degree of similarity of the script to a certain regional style or graphical mode. The profile consists of a pair of values for the graphical modes (square and cursive) and six regional weights on a scale from 0 to 10.

From the pages we extracted patches (square crops of the page image) of 350 x 350 pixels with a stride of 350. A patch was kept if its ink ratio under Otsu's thresholding [Otsu, 1979] fell in the interval [0.02; 0.60]. This condition eliminates blank margins and patches flooded by dark defects such as spots or stains, but it still admits patches near the page border and in blank areas where there is no writing (the trimmed edge, the binding, the imaging background). We call such patches text-free. Blind visual inspection of a sample of 120 patches by a single rater put their share at about one quarter (25%, rough 95% interval from 17% to 36%). In Section 5.1 and in the Appendix (A5) we check how they affect the results. The selected patches were resized to 224 x 224. To keep the training budget per page comparable, at most 60 patches were taken from each page, and the final sample contains 31,653 patches. Augmentation was limited to rotation by up to 3 degrees in either direction. Mirror flips were not used, since mirrored script does not occur in manuscripts. Each patch inherits the labels of its page, an assumption we discuss in Section 6.

The expert profile is mapped onto a distribution over the fourteen classes. Similar values for square and cursive indicate the semi-square mode [Droby et al., 2022b]. The pair of modes is expanded into square, semi-square and cursive by a piecewise linear scheme. The weight of the semi-square mode is set to twice the minimum of the original pair, and the remainder is divided between the pure modes. The products of the mode and regional weights are then normalized over the classes present in the dataset.

The expert profile is filled in for 714 of the 715 pages. After the mapping onto the fourteen classes, 690 of them still have a nonzero vector. For 24 Yemenite pages the entire mass falls on combinations that the dataset does not contain (see below). We checked the mapping on these 690 pages. The regional part is exact: the total variation distance between the expert distribution over the regional styles and the collapsed 14-dimensional vector does not exceed 0.037, the leading regional style does not change on any page, and nowhere does a secondary regional style gain 20 percentage points or more. We checked the mode part the same way. We collapsed the vector back into a pair of modes, splitting the semi-square weight evenly, and compared it with the normalized expert pair. For the script types that have all three modes in the dataset, the collapse returns the expert pair exactly. A discrepancy appears where the dataset lacks the required combination of regional style and mode, because the mass that falls on it is discarded by the normalization. It exceeds 0.05 only on 44 pages of Yemenite square script, where it reaches 0.300 (on 17 pages it is above 0.2) and the discarded share of the mass reaches 0.600. On Yemenite pages the expert assessment of mode hardly carries over into the 14-dimensional vector.

The expansion of the pair of modes into three modes remains an assumption. Of the 690 pages with a nonzero vector, the maximum of the mapped distribution disagrees with the hard label on 115 pages. On 71 of them the expert gave a mixed mode assessment (7/3, 4/6, 6/4 or 5/5) while the hard label was "square" or "cursive", and the scheme gives most of the mass to semi-square. In Section 5.3 we check which conclusions do not depend on this assumption: there the model is trained directly on the expert scores.

For 25 pages the mapping yields a zero vector. Twenty-four of them are labeled Yemenite square, yet the expert did not assign them a pure square mode: eighteen pages have mode 5/5 and six have 4/6. The entire mass falls on semi-square and cursive, and the dataset has no such Yemenite classes. One further page has an empty profile. These pages are excluded from every quantity that requires the expert vector, that is, from the JS divergence and from the set of transitional pages (hereafter rule M1). In accuracy, macro-F1 and ECE they take part on the same footing as the rest, since they do have a hard label. In training these pages stay in the sample, but under soft-label training they carry no information about the class. For cross-entropy with soft targets and for KL, the contribution of an all-zero target to the loss is zero; for mean squared error, a zero target pulls the output toward the uniform distribution. In Section 5.2 and in the Appendix (A7) we check whether the calibration gain of MSE holds on these pages.

## 4. METHOD

**4.1. Model and training.** We used ResNet50 [He et al., 2016] pretrained on ImageNet [Deng et al., 2009], with a new fully connected layer for fourteen classes. The models were trained with AdamW [Loshchilov, Hutter, 2019] (learning rate 3e-4, weight decay 1e-4) on a cosine schedule, for ten epochs, with batch size 64 (96 for ResNet18) and mixed precision. No validation set was held out: the number of epochs was fixed in advance and was the same for every configuration, and we used neither early stopping nor hyperparameter tuning. The random seed was held fixed across all main runs. For the soft target we repeated training with two further seeds, the hard model under S2 with seeds 0 and 1, plain MSE three more times (seeds 0, 1 and 2), and MSE without all-zero targets with those same three seeds. The original S2 runs were carried out on a Tesla P100, and the repeat runs and all runs with binarization on a Tesla T4 (Appendix, A4 and A6). To test the role of the background, we repeated protocols S1K, S2 and S3 (Section 4.4) on binarized patches: every patch was mapped to two values, ink and background, at an Otsu threshold computed for that patch, in training and at test time alike (in training, after augmentation). The folds, the patch filtering, the hyperparameters and the random seed were left unchanged. The S2 model on binarized patches was additionally trained with seed 1 (Section 5.1). To check whether the conclusions depend on the architecture, we repeated the protocol comparison with ResNet18 on the pair of protocols S1 and S2.

**4.2. Loss functions.** Hard-label training uses cross-entropy on the class label. For soft-label training we considered three variants: cross-entropy with soft targets, KL divergence, and mean squared error between the softmax output and the target distribution. The authors of the dataset used mean squared error for regression on the eight-dimensional expert vector (six regional styles and two modes) [Droby et al., 2022b]; we included this variant for comparability, but applied it to the distribution over the fourteen classes.

We also trained a separate variant directly on the expert scores, without the mapping onto the fourteen classes. It remains a single network, whose output layer is split into three groups of outputs. The fourteen classes are trained with hard cross-entropy. The six regional styles and the two modes are trained with soft cross-entropy applied directly to the expert scores. The regional scores and the mode scores are normalized independently of each other. The loss is the sum of three terms with equal weights, and examples with an all-zero target are excluded from the corresponding term. We trained this variant under protocol S2, on the same five folds and for the same ten epochs, and the results appear in Section 5.3.

**4.3. Aggregation of predictions.** The predicted distributions are averaged over the patches of a page, and the estimate for a manuscript is the average over all of its patches. All metrics are reported at three levels: patch, page and manuscript. Pages yield different numbers of patches, at most 60, so patch-level metrics are weighted by the amount of text on the page.

**4.4. Splitting protocols.** We compared three ways of splitting the data into a training and a test part, holding the model, the data and the training budget the same. Protocol S1K (naive) is page-wise cross-validation stratified by class: five folds, each page in the test set exactly once. As a result, pages of one manuscript appear in both the training and the test part. Protocol S2 (grouped) is manuscript-wise cross-validation with five folds (GroupKFold with the manuscript as the grouping variable). No manuscript is represented in both the training and the test part, and across the five folds every page of the dataset likewise enters the test set exactly once. Protocol S3 (official) reproduces the official split of the dataset into the training, typical test and blind test sets, and serves as an external point of comparison with the results published for the blind test set (Section 5.1). As a robustness check we added one more naive scheme, S1: five independent splits in a ratio of 80 to 20, whose pooled test set holds 482 distinct pages out of 715, some of them several times. S1 and S1K are different schemes. S1K is the main one, and S1 supports only the robustness check, the ResNet18 variant and the Grad-CAM maps (Section 5.4).

Our main measurement is the difference between the performance estimates under S1K and under S2. The two protocols differ in the variable by which the folds are formed, and that difference has two consequences that a design of this kind cannot separate. The training part of an S1K fold contains almost every manuscript of the dataset, from 169 to 171 of 171, while the training part of an S2 fold contains only 136 or 137. GroupKFold also does not stratify by class, so the test part of an S2 fold holds 12 or 13 of the 14 classes. We read the difference between the protocols as the combined effect of grouping by manuscript.

**4.5. Metrics and uncertainty estimation.** We measure classification performance with accuracy and macro-F1. Closeness of the predicted distribution to the expert profile is measured by the Jensen-Shannon divergence (JS) [Lin, 1991]. We compute it only on pages with a nonzero expert vector, under rule M1 from Section 3. This measure compares the weights the model gives to the fourteen classes with the expert scores for the same classes. It is zero when the two coincide exactly and grows as they diverge. Calibration is assessed by the expected calibration error (ECE) [Guo et al., 2017] over 15 equally spaced confidence bins. At the manuscript level there are only 171 units, i.e. the entire dataset. For the difference between protocols that is enough (Section 5.1), but for differences of a few percentage points between training schemes it is not (Section 5.2). For ECE it is too few. To gauge the noise, we drew the correctness of each prediction 10,000 times as a random event with probability equal to the model's confidence, which is how a perfectly calibrated model behaves, and each time we recomputed ECE over the same 15 bins. Even for such a model, manuscript-level ECE averages about 0.09 (95th percentile 0.12) and page-level ECE about 0.04. So we draw our conclusions about calibration at the patch and page levels, and report manuscript-level ECE for reference only.

Confidence intervals (CI) were obtained by a percentile bootstrap with resampling at the manuscript level [Field, Welsh, 2007], with 10,000 replicates at the 95% level. Resampling individual observations is unsuitable here, because it would reproduce the leakage inside the interval. Differences between protocols and between training schemes were estimated with a paired cluster bootstrap: the comparison runs on the same test units, manuscripts are resampled, and the number of replicates and the confidence level are the same. We call a difference statistically significant when its 95% bootstrap interval excludes zero. We treat an interval as excluding zero only if this still holds after its bounds are rounded to three decimals. The bootstrap was carried out with the fold models held fixed. It describes the variability that comes from the composition of the test set. It covers neither the run-to-run variation nor the dependence between folds whose training sets overlap. Therefore, the intervals we report are a lower bound on the uncertainty; run-to-run variation is estimated separately in Section 5.2. No correction for multiple comparisons was applied, and each interval describes its own comparison.

We call a page transitional when the maximum of the mapped expert distribution lies strictly between 0 and 0.6, so that no class receives a confident majority. There are 97 such pages. Pages with an all-zero vector are not among them, by rule M1. In this dataset the expert profile is defined essentially per manuscript, and for 161 of the 171 manuscripts it is the same on every page. So being transitional is a property of the manuscript, and the 97 transitional pages come from only 16 manuscripts. A bootstrap over manuscripts is unreliable with so few clusters. On the transitional subsamples we use an exact manuscript-level permutation test instead: the per-manuscript sums of the differences are sign-flipped in all possible ways, and the test statistic is again the difference of page-level means. The distribution of the maximum is discrete, and 87 pages sit exactly at 0.6. The composition of the subsample is thus sensitive to the threshold, and the Appendix (Table A2) reports a check at thresholds from 0.5 to 0.65.

# 5. EXPERIMENTS

## 5.1. The cost of manuscript-level leakage.

Table 1. Accuracy under the two splitting protocols (ResNet50, five folds). Square brackets give 95% confidence intervals from a cluster bootstrap over manuscripts. The upper rows were obtained on grayscale patches, the lower three on patches binarized with Otsu's method (the run with random seed 0 on a Tesla T4, Section 4.1).

| protocol | patch | page | manuscript |
|---|---|---|---|
| S1K (page-wise) | 0.896 [0.872; 0.916] | 0.972 [0.953; 0.988] | 0.994 [0.982; 1.000] |
| S2 (manuscript-wise) | 0.430 [0.371; 0.493] | 0.541 [0.470; 0.614] | 0.591 [0.520; 0.667] |
| difference S1K minus S2, percentage points | +46.6 [+40.8; +52.2] | +43.1 [+35.9; +50.1] | +40.4 [+33.3; +47.4] |
| S1K, binary input | 0.795 [0.771; 0.817] | 0.975 [0.956; 0.990] | 0.994 [0.982; 1.000] |
| S2, binary input | 0.441 [0.384; 0.499] | 0.582 [0.509; 0.657] | 0.655 [0.585; 0.725] |
| difference S1K minus S2 with binary input, percentage points | +35.5 [+30.2; +40.7] | +39.3 [+32.1; +46.4] | +33.9 [+26.9; +40.9] |

The difference between the protocols is 46.6, 43.1 and 40.4 percentage points at the patch, page and manuscript levels, and none of the intervals contains zero. In both protocols every page enters the test set exactly once, so at the page and manuscript levels the difference is paired by construction. It is computed on the same 715 pages and the same 171 manuscripts.

The additional naive scheme S1 (five 80 to 20 splits) gives similar values: 45.0, 42.3 and 39.2 pp. For ResNet18 we computed the pair S1 and S2, and there the difference is 45.1, 45.6 and 40.3 pp. Repeated runs of the S2 model, the removal of text-free patches and binarization of the patches also change the page-level gap only slightly. In all of these variants it runs from 39 to 44 pp, and no interval contains zero (Figure 1; Appendix, A4 to A6). On the patches flagged as text-free by the automatic criterion, accuracy is 0.728 for the S1K model and 0.246 for the S2 model. The model recognizes a familiar manuscript even where there is no writing, that is, from the writing support or the imaging conditions (A5).

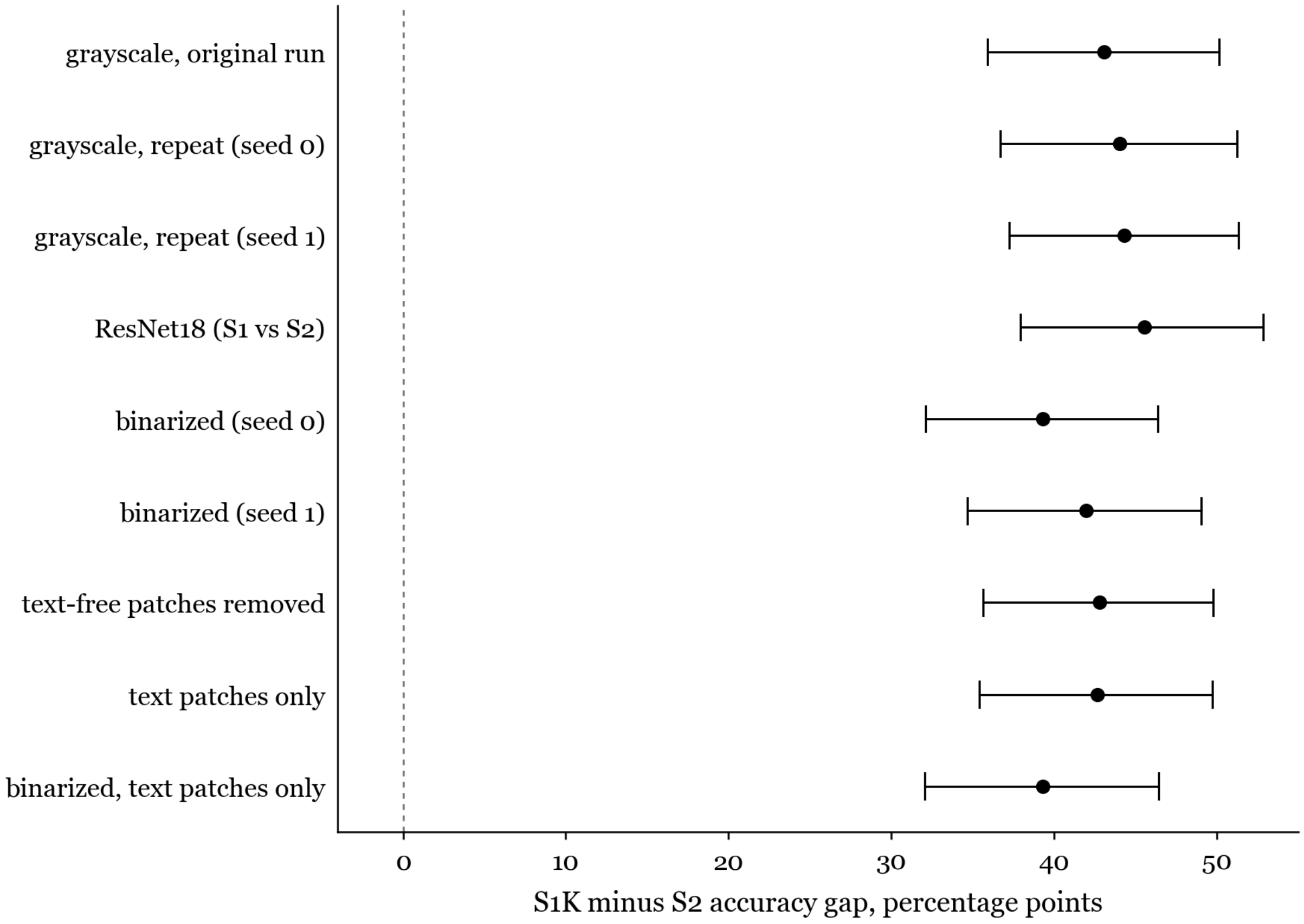

Figure 1. The gap between the two protocols across variants of the experiment. Page-level accuracy of S1K minus S2, in percentage points, with 95% confidence intervals from a cluster bootstrap over manuscripts (10,000 replicates). The ResNet18 row compares S1 with S2, because no S1K run was made for that backbone. The dashed line marks zero.

Table 2. The official split of the dataset (protocol S3, ResNet50): accuracy on the typical and blind test sets.

| test set | patch | page | manuscript |
|---|---|---|---|
| typical (the same manuscripts as in training) | 0.915 [0.880; 0.943] | 0.972 [0.942; 0.993] | 0.969 [0.938; 0.992] |
| blind (unseen manuscripts) | 0.334 [0.227; 0.448] | 0.390 [0.256; 0.528] | 0.439 [0.293; 0.585] |

The official test sets of the dataset agree with this picture. The estimate on the typical test set matches S1K: at the page level both are 0.972. The S2 estimate (0.541) lies between S1K and the official blind set (0.390). The difficulty of the blind manuscripts themselves does not explain this difference: if the S2 estimate is restricted to the same 41 manuscripts that make up the blind set, it becomes 0.576 [0.431; 0.732]. The protocols differ in what their training part contains. Under S2 the model is trained on roughly 570 pages, which include other manuscripts from this group as well as manuscripts of the same script types. The official split leaves 400 training pages and removes the whole group of 41 manuscripts at once. So the S2 estimate is probably optimistic too for entirely new material.

Our absolute accuracy on the blind test set is lower than the accuracy published by the authors of the dataset. In their pipeline ResNet50 classifies 52% of the blind patches correctly with grayscale input and 62% with binary input, and their best models reach 70.4% (VGG19) and 75% (the hierarchical model) at the page level [Droby et al., 2022a]; our ResNet50 with grayscale input reaches 0.334 on patches and 0.390 on pages of the same set. Our pipelines differ. They extract patches with a uniform text scale (on average five lines in each patch) and train the network on 50,000 patches, first for 20 epochs with a frozen backbone (the pretrained weights are kept unchanged) and then for another 20 epochs with the whole network trainable. They assign the page the majority class of nine randomly sampled patches, and they consider binarized patches the best input. We use patches of a fixed size without overlap, ten epochs, averaging of the predicted distributions over all patches of a page, and grayscale input. We compare the protocols within a single pipeline, and with a different pipeline the size of the gap may be different. The authors attribute the gain from binarization to the removal of the background and to reduced overfitting. If the model recognizes a familiar manuscript by its background, the gap should narrow after binarization, so we repeated S1K, S2 and S3 on binarized patches (Section 4.1). The interpretation rule was fixed before the results were available.

With binary input the difference S1K minus S2 at the page level is +39.3 pp [+32.1; +46.4] (Table 1), and in the second run of the binarized S2 model it is +42.0 pp [+34.7; +49.0]. Under that rule (a gap of at least 30 pp with an interval that excludes zero) the main conclusion holds: the leakage does not reduce to the background and support cues that binarization removes. The accuracy gain from binarization on unseen manuscripts is not established, and Otsu binarization in our pipeline does not close the shortfall relative to the numbers reported by the authors of the dataset (Appendix, A4).

Calibration depends on the protocol as well. Under S1K the patch-level ECE is 0.065; under S2 it rises to 0.444. Figure 2 shows where that difference comes from. The paired page-level difference in ECE (S1K minus S2) is -0.106 [-0.184; -0.054]. Under evaluation with leakage the model looks calibrated, whereas on unseen manuscripts it keeps a high confidence even where it is wrong. At the page level, mean confidence under the manuscript-level split is 0.70 against an accuracy of 0.54 (under the page-level split, 0.92 against 0.97). If the pages are ordered by confidence, 67% of the predictions are correct in the more confident half and 41% in the less confident half.

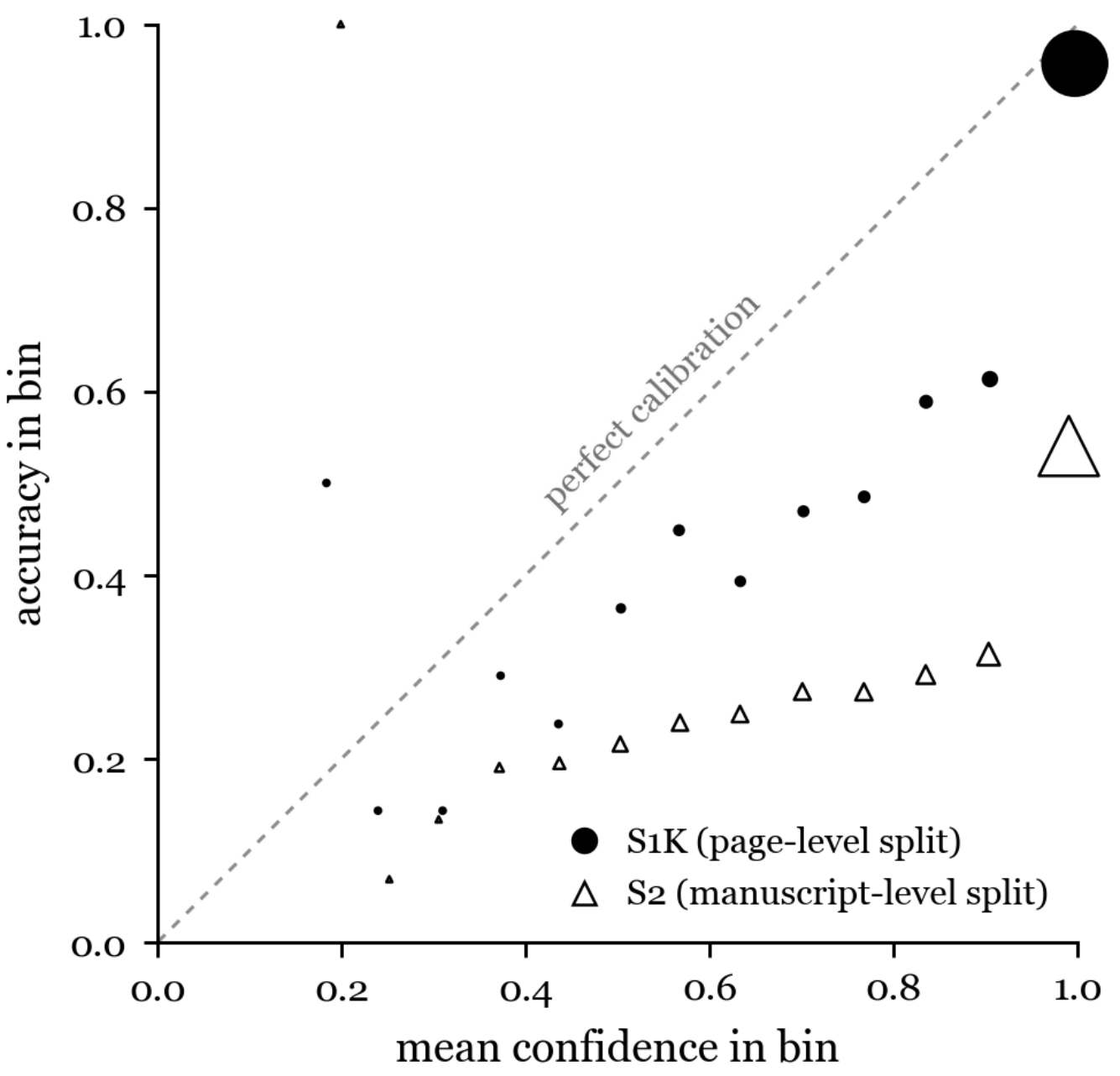

Figure 2. Reliability diagram at the patch level for hard cross-entropy, over 15 equally spaced confidence bins. Marker area is proportional to the number of patches in the bin. The top bin holds 27,503 of the 31,653 patches under S1K and 19,034 under S2. At almost the same confidence there, 0.997 against 0.990, accuracy is 0.957 against 0.540. The sparse bins near the left edge hold one or two patches each.

With binary input the overconfidence on unseen manuscripts remains at the patch level, with mean confidence 0.831 against an accuracy of 0.441, while at the page level it is markedly smaller: ECE of 0.063 against 0.172 for the grayscale model trained on the same hardware with the same random seed (paired difference -0.109 [-0.157; -0.021]). We attribute this reduction to binarization together with the poorer fit of the binarized models to the training patches. If the text-free patches are left out, the overconfidence is larger (Appendix, A4).

## 5.2. Hard and soft labels.

Table 3. Training variants (protocol S2, ResNet50, five folds). Values separated by slashes are patch / page / manuscript. The last row belongs to the variant from Section 5.3. For hard CE (the original run and two repeats) and for MSE without all-zero targets we also give the range across three runs (Appendix, A6 and A7).

| training | accuracy | macro-F1 | JS | ECE |
|---|---|---|---|---|
| hard CE | 0.430 / 0.541 / 0.591 | 0.385 / 0.501 / 0.515 | 0.351 / 0.290 / 0.267 | 0.444 / 0.163 / 0.143 |
| hard CE, 3 runs | 0.430-0.439 / 0.529-0.541 / 0.585-0.602 | 0.385-0.400 / 0.489-0.501 / 0.515-0.537 | 0.351-0.352 / 0.290-0.294 / 0.267-0.272 | 0.436-0.444 / 0.163-0.172 / 0.125-0.143 |
| soft CE | 0.427 / 0.509 / 0.579 | 0.373 / 0.447 / 0.473 | 0.300 / 0.253 / 0.238 | 0.296 / 0.094 / 0.088 |
| KL | 0.423 / 0.522 / 0.626 | 0.365 / 0.456 / 0.507 | 0.297 / 0.251 / 0.236 | 0.298 / 0.082 / 0.112 |
| MSE | 0.430 / 0.519 / 0.591 | 0.377 / 0.458 / 0.472 | 0.294 / 0.256 / 0.244 | 0.209 / 0.052 / 0.071 |
| MSE without all-zero targets, 3 runs | 0.423-0.427 / 0.515-0.526 / 0.591-0.632 | 0.371-0.374 / 0.457-0.465 / 0.479-0.512 | 0.291-0.294 / 0.250-0.256 / 0.238-0.245 | 0.240-0.244 / 0.057-0.069 / 0.076-0.097 |
| label smoothing (epsilon = 0.1) | 0.443 / 0.520 / 0.591 | 0.396 / 0.487 / 0.515 | 0.338 / 0.300 / 0.284 | 0.269 / 0.092 / 0.095 |
| direct variant, class outputs | 0.432 / 0.515 / 0.561 | 0.390 / 0.483 / 0.487 | 0.345 / 0.285 / 0.263 | 0.436 / 0.178 / 0.122 |

Confidence intervals for page-level accuracy are [0.470; 0.614] for hard CE, [0.436; 0.585] for soft CE, [0.447; 0.599] for KL, [0.447; 0.596] for MSE and [0.447; 0.596] for label smoothing.

Comparing these intervals by eye is not enough, so all differences were computed in paired form, on the same test units of protocol S2 (Table A1 in the Appendix).

For accuracy, no difference from hard cross-entropy is significant, either at the page level or at the manuscript level. The intervals are wide, from 6 to 14 percentage points: the data are compatible both with no difference and with a difference of up to 9 or 10 pp. We found no effect, but these data do not establish equality of accuracy either. In JS divergence all three soft variants are significantly closer to the expert profile, and this holds at both aggregation levels. For plain MSE at the manuscript level it holds in the run reported in Table 3, but not in all of the repeat runs (A6). The calibration gain is significant at the patch level, from -0.146 [-0.188; -0.105] for KL to -0.235 [-0.275; -0.195] for MSE, and at the page level, from -0.069 to -0.111. At the manuscript level the intervals no longer exclude zero.

The divergence from the expert profile has the same form as what the soft training variants optimize, so a comparison on it is not neutral. The result is still informative, for two reasons. First, the divergence is measured on manuscripts the model has not seen. Second, label smoothing, which also blurs the target, gives no gain on this metric (see below).

In absolute terms, the JS divergence on patches drops from 0.351 under hard training to between 0.294 and 0.300 under soft training, and on pages from 0.290 to between 0.251 and 0.256. The calibration error on patches falls from 0.444 to between 0.209 and 0.298, and on pages from 0.163 to between 0.052 and 0.094 (by Table 3). With the repeat runs included, the two ranges are 0.209 to 0.308 and 0.052 to 0.112.

Table 4. Transitional pages (the maximum of the mapped distribution lies between zero and 0.6), 97 pages from 16 manuscripts, page level. The differences are paired against hard cross-entropy. In parentheses, the p value of the exact permutation test over manuscripts. pp = percentage points.

| training | accuracy | JS | difference in accuracy, pp (p) | difference in JS (p) |
|---|---|---|---|---|
| hard CE | 0.443 | 0.386 | | |
| soft CE | 0.247 | 0.275 | -19.6 (0.19) | -0.111 (0.010) |
| KL | 0.237 | 0.274 | -20.6 (0.19) | -0.112 (0.005) |
| MSE | 0.278 | 0.271 | -16.5 (0.19) | -0.115 (0.006) |
| label smoothing (epsilon = 0.1) | 0.433 | 0.351 | -1.0 (1.00) | -0.035 (0.12) |

Of the 97 transitional pages, 36 are transitional by the expert scores for the regional styles, while the remaining 61 fall under the threshold because of a mixed score for the graphical mode, that is, through the assumption about the expansion of the mode scores into three modes (Section 3). On transitional pages the gap in divergence is wider than on the full dataset, 0.386 under hard training against 0.271 to 0.275 under soft training. All three differences are significant under the permutation test as well (p from 0.005 to 0.010). In the repeat runs with the soft target one of the two is significant (p = 0.036 and 0.12).

Argmax accuracy on the same pages is higher for hard training, 0.443 against 0.237 to 0.278, with differences from -16.5 to -20.6 pp. Statistically this loss is not confirmed (p = 0.19). Accuracy changes in only five of the sixteen manuscripts, and with that few of them the permutation test has little sensitivity (the smallest attainable p value is 0.0625). In the repeat runs with the soft target the difference is -20.6 and -23.7 pp (p = 0.094 and 0.0625). In the second of them all five manuscripts whose accuracy changed lose accuracy. The hard label on these pages is conventional: the expert gave no single class a preference, while an argmax metric requires one to be chosen. If a single label is needed from the model, the possible loss of accuracy on such manuscripts should be checked separately. Where this threshold is set changes the conclusion quantitatively but not in sign (Appendix, Table A2).

A label-smoothing ablation [Szegedy et al., 2016] (epsilon = 0.1) marks out the limits of the cheap alternative. In patch-level accuracy and macro-F1 this variant turns out to be the best in Table 3, and it improves calibration almost as much as the expert labels (patch-level ECE 0.269 against 0.444 for hard training). Smoothing does not give closeness to the expert profile. Its page-level JS is 0.300, against 0.290 for hard training and 0.256 for the MSE variant. We found no evidence of a difference from hard training on this metric at the page level (+0.011 [-0.001; +0.023]). At the manuscript level smoothing is worse than hard training (+0.017 [+0.006; +0.029]), in two of the three comparisons with the runs of the hard model, and MSE beats smoothing by -0.044 [-0.065; -0.025]. On transitional pages the gap is larger, 0.351 against 0.271. Smoothing removes the excess confidence. It carries nothing about which script types are close to one another, and that information is only in the expert labeling.

Calibration after training is usually corrected by temperature scaling: the model's logits are divided by a single scalar T, tuned by minimizing the negative log likelihood of the true labels (NLL). At the page and manuscript levels we tested two options: in the first, a T tuned on patches divides the patch logits before averaging, and in the second, a separate T is tuned on the averaged distributions of that level and divides their logarithm. With T greater than 1 the confidence goes down, while the prediction on a patch does not change. We tested scaling on the hard model. The protocol has no validation set, so T was tuned inside the test fold, on other manuscripts. The test manuscripts of a fold were split into five subgroups, and for each subgroup T was tuned on the remaining four. The model had seen none of these manuscripts in training, and the manuscripts on which T was tuned for a subgroup do not enter the evaluation of that subgroup. For this calculation the logits of the hard model were recomputed from its saved weights. The prediction files store probabilities to five decimals, and tuning T by likelihood would depend on that rounding. On patches, scaling calibrates better than soft labels. ECE falls from 0.444 to 0.064, against 0.209 to 0.308 for the soft variants, and the paired difference with each of them is significant.

Once the predictions are averaged over the page and over the manuscript, the gain disappears: ECE is 0.158 on pages (-0.008 [-0.129; +0.103]) and 0.230 on manuscripts (+0.090 [-0.032; +0.163]), against 0.166 and 0.140 without scaling. These two values were computed from the recomputed logits, whereas Table 3 gives 0.163 and 0.143. The original predictions ran on a GPU in half precision and the repeats on a CPU in full precision (argmax agrees on 99.74% of patches). The correction these levels need is a different one: a temperature tuned directly on the averaged distributions of the same level falls from between 3.8 and 5.4 on patches to between 2.0 and 3.4 on pages and between 1.4 and 2.3 on manuscripts (medians across folds). It too gives no significant improvement: 0.163 on pages (-0.003 [-0.103; +0.093]) and 0.137 on manuscripts (-0.003 [-0.079; +0.081]). A temperature tuned by NLL turns overconfidence into underconfidence. After scaling, mean confidence on pages is 0.436 against an accuracy of 0.541, and on manuscripts 0.509 against 0.591. We tested only scalar scaling (one temperature for all classes, tuned by NLL), in the two ways of applying it.

For the soft variants, the page-level calibration gain relative to the hard model without scaling is significant for MSE (in the run of Table 3 and in all three runs without all-zero targets) and in three of the four runs with the soft target. ECE is 0.052 to 0.112 against 0.163 for the original hard model (the comparison with its three runs is in the Appendix, A6). At the manuscript level ECE is 0.071 to 0.112 against 0.143, but the intervals of the differences include zero (Table A1).

Cross-entropy with soft targets and the KL divergence differ by a term that does not depend on the model parameters, so their gradients coincide. A check confirms this: with the same seed of the random number generator, the two variants give fully matching predictions (two repeat runs of each variant, carried out independently, with identical prediction files). The "soft CE" and "KL" rows in Tables 3 and A1 were obtained earlier, on different hardware (Tesla P100, while the repeat runs ran on a Tesla T4). On that device, training with the same seed did not reproduce exactly. They should be read as two runs of one and the same procedure. The accuracy difference between them at the manuscript level is +4.7 pp [+1.2; +8.8], although only the run differs.

The soft target has four distinguishable training runs in all, the two original ones and two repeats with different seeds, with the same folds and the same patch filtering. Manuscript-level accuracy varies across them from 0.573 to 0.626 (a range of 5.3 pp), and page-level accuracy from 0.499 to 0.522 (2.2 pp). The difference of +4.7 pp quoted above refers to the two original runs. The largest difference, +5.3 pp [+0.6; +9.9], is between the original KL run and the repeat with seed 2. Divergence and calibration are more stable: page-level JS stays between 0.251 and 0.259, and page-level ECE between 0.082 and 0.112. Run-to-run variation does not affect the conclusion about closeness to the expert. The JS difference from the original hard model is significant in each of the four runs at both levels, and on pages it runs from -0.031 to -0.038, roughly four times the range across runs. The page-level calibration gain against the original hard model is significant in three runs out of four. For accuracy the conclusion is different: differences between training variants smaller than about 5 pp at the manuscript level cannot be established from a single run, and the intervals computed on the test set do not show them. In two pairs out of six, the bootstrap interval of the difference between runs of the same target does not contain zero.

For the hard model, repeated on a Tesla T4 with seeds 0 and 1, the variation between runs is smaller: 1.8 pp at the manuscript level against 5.3 pp for the soft target, although with that number of runs (three and four) the difference is not established. The conclusions about divergence and about page-level calibration also hold when each of the four runs with the soft target is compared with each of the three runs of the hard model (Appendix, A6). We found no evidence of an accuracy difference in any pair, but at the page level all four runs with the soft target are below all three runs of the hard model, 0.509 against 0.534 on average. A permutation test over runs gives p = 2/35, about 0.06, so a small loss of accuracy at the page level cannot be ruled out.

The calibration gain of MSE is not explained by the pages with an all-zero vector, whose all-zero target for MSE pulls the output toward the uniform distribution (Section 3). When the all-zero targets are excluded from the loss (three runs), the difference in ECE from hard training remains significant at the patch level and at the page level. On patches, ECE without the all-zero targets is slightly higher than for plain MSE (+0.020), and such a small degradation cannot be ruled out (Appendix, A7).

**5.3. Training on expert scores without the mapping.** Section 4.2 describes the scheme of this variant: the expert scores for the regional styles and the modes enter the loss directly. To compare against the original expert numbers, we collapsed the predictions of the other variants onto the six regional styles and the two modes. The mass of each semi-square class is split equally between the square and the cursive mode. Predictions here are aggregated at the page level. The reference is the original expert scores, without the mapping. The profile is filled in for 714 pages, so the comparison also brings back the 24 Yemenite pages for which the mapping gives an all-zero vector.

Table 5. Closeness to the expert scores without the mapping (protocol S2, page level). All 714 pages, and the 97 transitional pages (16 manuscripts) from Section 5.2.

| training | JS, regional style | JS, mode | top-1, regional style | JS, regional style, transitional | JS, mode, transitional |
|---|---|---|---|---|---|
| hard CE | 0.201 | 0.040 | 0.679 | 0.290 | 0.026 |
| soft CE | 0.193 | 0.033 | 0.651 | 0.196 | 0.026 |
| KL | 0.189 | 0.033 | 0.679 | 0.188 | 0.026 |
| MSE | 0.200 | 0.030 | 0.667 | 0.204 | 0.022 |
| label smoothing (epsilon = 0.1) | 0.216 | 0.035 | 0.657 | 0.264 | 0.020 |
| direct outputs, regional style and mode | 0.191 | 0.025 | 0.675 | 0.223 | 0.021 |

For the regional styles we found no evidence of a difference between the direct outputs and cross-entropy with soft fourteen-class labels. The paired JS difference is -0.001 [-0.013; +0.009] (Table A3 in the Appendix). The direct outputs are significantly closer to the expert than label smoothing is: -0.025 [-0.039; -0.012]. For the mode they are closer to the expert than any other variant, with JS 0.025 against values from 0.030 to 0.040. All five paired differences for the mode are significant, from -0.005 [-0.010; -0.001] against MSE to -0.015 [-0.022; -0.007] against hard training. For top-1 regional style we found no evidence of a difference either: 0.675 for the direct outputs and 0.679 for hard training, a paired difference of -0.4 percentage points [-4.4; +3.5]. On the full dataset the variants differ little for the regional styles (JS from 0.189 to 0.216), and between the direct outputs and hard training we found no evidence of a difference (-0.009 [-0.026; +0.006]).

On the transitional pages (97 pages, 16 manuscripts) the gain from soft labeling is visible in the space of the expert scores as well. It concerns the regional style: JS for the regional style drops from 0.290 under hard training to between 0.188 and 0.204 for the soft variants (manuscript-level permutation test: p from 0.013 to 0.024; in the repeated runs of the soft target p = 0.21 and 0.041), while for the mode we found no evidence of a difference. For the direct outputs, JS for the regional style on these pages is 0.223. The difference from hard training has the same sign, -0.066, but it is not significant (p = 0.081). The direct outputs trail soft cross-entropy by +0.028, and this difference is not significant either (p = 0.079). In the repeated runs of the soft target it is -0.006 and +0.038. These data do not allow us to establish which of the two formulations works better on transitional pages.

The direct variant differs from soft training in more than dropping the mapping: it also changes how the expert assessment is represented and how the output layer is built. We therefore compare the two designs as a whole, and the difference cannot be attributed to dropping the mapping alone. In addition, the set of transitional pages is itself selected through the mapping. Selecting them without it, by the expert scores for the regional styles (maximum score below 0.6), gives 36 pages from six manuscripts, and all of them are among the 97. With six manuscripts a permutation test cannot return a p value below 0.031, so we report this check as descriptive. On it, JS for the regional style is 0.304 under hard training, between 0.190 and 0.206 for the soft variants and 0.270 for the direct outputs. The reduction for the soft variants comes mostly from three of the six manuscripts.

This series does not test the claims of Section 5.2 about accuracy and about calibration. The class outputs of the direct variant are trained with hard cross-entropy, and their calibration is close to that of hard training: ECE 0.436 / 0.178 / 0.122 against 0.444 / 0.163 / 0.143 (Table 3). This is consistent with the calibration gain in Section 5.2 being associated with the soft target on the classes themselves. For accuracy, the class outputs give 0.515 at the page level against 0.541 for pure hard training; the paired difference is -2.7 percentage points [-6.9; +1.3] at the page level and -2.9 percentage points [-8.2; +1.8] at the manuscript level. We found no evidence of a significant loss of accuracy from sharing one network across three tasks.

**5.4. Qualitative analysis.** For the models trained under protocols S1 and S2 we computed Grad-CAM maps [Selvaraju et al., 2017]. Like the ResNet18 check, this analysis uses the additional scheme S1. The maps were computed on pages that neither model had seen in training. These are the 37 pages that the test set of the first fold has in common under both schemes, drawn from 22 manuscripts, over all 1,575 patches of those pages. The leakage shows up on them in accuracy: the S1 model, which had seen other pages of 21 of these 22 manuscripts, is correct on 0.816 of the patches, the S2 model on 0.363. The fraction of Grad-CAM mass inside the ink mask is nevertheless almost the same for the two models, 0.264 and 0.259, a difference of +0.005 [-0.009; +0.020] (manuscript-level permutation test: p = 0.74). This proxy measure does not identify the source of the leakage. On text-free patches the S1K model still recognizes a familiar manuscript (Section 5.1), but on patches that almost always do contain writing the patch-level gap is nearly the same as on all patches (+45.7 percentage points [+39.2; +51.8] against +46.6), so the leakage does not reduce to text-free patches. The map from the last convolutional layer (7 x 7) is a coarse localization map relative to the 224 x 224 patch, so we draw no conclusions about the source of the leakage from the maps.

## 6. DISCUSSION AND RECOMMENDATIONS

An accuracy figure reported in a digital paleography paper should be read together with the description of how the model was tested. Two things matter. Were pages of the same manuscript used both for training and for testing? And what counted as one answer: a patch (a square crop of the page image), a page, or a whole manuscript? Testing on unseen pages of the manuscripts used for training gives 97% of pages correct on our dataset. The model does not rely on the writing alone. On patches near the page border, where there is no writing, it names the correct class in 73% of cases if it has seen other pages of the same manuscript, and in 25% if it has not. On a previously unseen manuscript about half the attributions are correct, or fewer: keeping the manuscripts used for training and for testing apart gives 54% of pages and 59% of manuscripts, and the official set of unseen manuscripts gives 39% of pages. The model's confidence there is too high, 70% on average when 54% are correct, and it cannot be read as the probability that the attribution is correct. If the images are binarized, that is, mapped to two values, ink and background, the overconfidence becomes noticeably smaller at the page level, but at the patch level it remains. In the more confident half of the pages 67% of the attributions are correct, in the less confident half 41%, so it is better to start re-examination with the less confident ones. About a third of the attributions in the more confident half are wrong too. For manuscripts that the expert assigns to several script types at once, a model trained on soft labels comes closer to the expert's distribution, but by the point estimate its single answer on pages of such manuscripts agrees with the hard label less often (0.21 to 0.28 against 0.44 on the 97 transitional pages); across the 16 manuscripts this difference is not established.

1. Treat a grouped split at the manuscript level as the minimum requirement: on our dataset a page-level split inflates accuracy by 40 to 47 percentage points, and by 34 to 42 with binarized patches.

2. Always state the level at which predictions are aggregated (patch, page or manuscript), because the number itself depends on it: under the manuscript-level split, accuracy rises from patch to manuscript, from 0.430 to 0.591.

3. Report calibration alongside accuracy, and report it at the aggregation level at which the predictions will be used. On unseen manuscripts the model's confidence no longer matches its accuracy (0.70 against 0.54 at the page level), although it is still useful for deciding which attributions to re-examine: 67% of the predictions are correct in the more confident half of the pages and 41% in the less confident half. Also check calibration on the kind of input images the model will work with: with binarized patches, page-level overconfidence is noticeably smaller, while at the patch level it remains.

4. For datasets with transitional script types, it is worth reporting the distance to the expert distribution alongside accuracy.

5. Report differences between training variants in paired form, with a confidence interval on the difference itself. Overlap between the separate intervals answers a different question and is easily misleading.

6. Estimate run-to-run variation separately. Intervals computed on the test set do not cover it: for the soft target, the difference between runs of one and the same procedure reaches 5.3 percentage points at the manuscript level (Section 5.2), and the bootstrap calls it significant. For the hard model it reaches 1.8 percentage points.

7. When the expert annotation is given per manuscript, check conclusions about subgroups of pages, transitional pages for example, against the number of manuscripts rather than the number of pages: the 97 transitional pages of our dataset come from 16 manuscripts.

We used a single dataset. Other public datasets did not fit a grouped protocol: in Historical-WI the grouping variable coincides with the class (the scribe); in the training part of the CLaMM dataset [Cloppet et al., 2017] the manuscript shelfmark is missing for 3,522 images out of 3,540; in tasks 2 and 4 of the same competition only 39 manuscripts have two or more pages (131 images). We have one external point of comparison, the official split of the dataset, and it is stricter than our grouped protocol. On the same 41 manuscripts S2 gives 0.576 and the official blind set 0.390. We did not test grouping by collection or by digitization campaign; if the model relies on imaging conditions, the S2 estimate may be optimistic. Each patch inherits the label of its page.

We used two variants of an architecture from one family. We held out no validation set, so no configuration was brought to its own best accuracy. For the same reason we tuned the temperature for scaling inside the test folds, on other manuscripts, and we tested only scalar scaling (one temperature for all classes, tuned by NLL). We measured run-to-run variation for the hard model under S2 (three runs), for the soft target (four) and for MSE (four runs of plain MSE and three without all-zero targets, Section 5.2 and Appendix, A6 and A7). The remaining configurations are represented by a single run each: S1K, S1, S3, ResNet18, label smoothing, the direct variant, and, on binarized patches, S1K and S3. S2 on binarized patches is represented by two runs. The repeated runs were not carried out on the same hardware as the original ones, so a change of hardware is part of their variation (Tesla P100 against Tesla T4). With the random seed held the same, this change moved the accuracy of the hard model by 1.0 percentage point at the page level and by 0.6 percentage points at the manuscript level. Neither difference is significant.

The expansion of the pair of mode scores into three graphical modes remains an assumption, and for the Yemenite pages the mapping loses the expert's assessment of the mode (Section 3). Closeness to the expert profile is measured by a quantity of the same form as the one soft-label training optimizes; we have no measure of agreement with the expert that is neutral with respect to the training scheme. The expert profile is given essentially per manuscript, so the conclusions about transitional pages rest on 16 manuscripts, on seven at a threshold of 0.5, and on six when the pages are selected without the mapping. The gain from soft labeling on transitional pages depends quantitatively on the threshold, and by the point estimate it is weaker when the model is trained directly on the expert scores, whereas the loss of accuracy on these pages (point estimate 16.5 to 23.7 percentage points) can neither be confirmed nor ruled out with 16 manuscripts.

We did not isolate the causal mechanism of the leakage. Otsu's thresholding does not close the gap (Section 5.1), so the leakage does not reduce to the background and writing-support cues that binarization removes. What survives binarization is the hand of a particular scribe, the size of the letters and the spacing between the lines, and, on patches near the page border, the outlines of the trimmed edge, the binding and the imaging background; we did not test which of these cues the model uses. About a quarter of the patches contain no writing (Section 3). They barely affect the main gap, but the model recognizes familiar manuscripts from them, and page-level calibration depends on them (Section 5.1; Appendix, A4 and A5). We did not test training on patches selected more strictly. In other studies suppressing the background improved results on unseen manuscripts: background filtering on the blind set of VML-HP [Droby et al., 2021] and binarization on the blind set of VML-HP-ext [Droby et al., 2022a]. On grayscale images of the codices of Test B, background masking: the mean F1 of three networks is 0.38 without the mask and 0.47 with it [Weißmann et al., 2024]. In our two binarized runs, page-level S2 accuracy is 1.4 to 5.3 percentage points higher than in the three grayscale ones, but the difference is significant in only two of the six comparisons, so the gain is not established. Page-level calibration depends on whether the model receives grayscale or binarized images (Section 5.1); we tested soft labels and temperature scaling only on grayscale input, and scaling only on the hard model. The evaluation protocol (a grouped split, a stated aggregation level, calibration and paired comparisons) carries over to any dataset in which the manuscript is known for every image and every class is represented by several manuscripts. The size of the effects on other material, Slavonic manuscripts for example, has to be measured anew.

## 7. CONCLUSION

How the data are split moves the reported accuracy more than the choice of architecture does. With the model, the dataset and the training budget held fixed, testing on other pages of manuscripts the model has already seen puts page-level accuracy at 97%. Testing on manuscripts it has not seen puts it at 54%, and the official test set of unseen manuscripts at 39%. The gap of 40 to 47 percentage points held up in every check we ran: on a second architecture, in repeated training runs, on binarized patches (from 34 to 42 percentage points) and with the patches that carry no writing removed. Part of the mechanism is visible, though not all of it. On patches without writing, a model that has seen other pages of the same manuscript gives the correct class in 73% of cases, against 25% for a model that has not seen that manuscript.

A published accuracy figure says nothing about how a model will behave on a new manuscript unless the split and the aggregation level (patch, page, manuscript) are stated with it. On unseen material the model's confidence is systematically too high, so it cannot be used as the probability that a prediction is correct, but it can still order the pages for re-examination. Expert class scores, where a dataset provides them, bring the prediction closer to the expert's assessment and lower the calibration error, but they add no accuracy, and our data do not rule out a small loss of it. At a minimum, the data should be split by manuscript, and a report should name the level at which the metrics were computed.

What exactly gives a manuscript away after binarization we did not establish. Nor did we find the reason our pipeline is less accurate in absolute terms than the authors of the dataset report. Binarization does not explain it. The code will be released together with the paper.

## APPENDIX

A1. Paired differences between training variants. They are discussed in Section 5.2.

Table A1. Paired differences between training variants (protocol S2; cluster bootstrap over manuscripts, 10,000 replicates, 95%; asterisks mark intervals that exclude zero). For MSE without all-zero targets we take the run with random seed 0, as in the other rows. The two other runs are reported in A7.

| comparison | level | accuracy | JS | ECE |
|---|---|---|---|---|
| soft CE minus hard CE | page | -0.032 [-0.088; +0.023] | -0.036 [-0.061; -0.012]* | -0.069 [-0.126; -0.016]* |
| KL minus hard CE | page | -0.020 [-0.079; +0.038] | -0.038 [-0.063; -0.014]* | -0.081 [-0.132; -0.018]* |
| MSE minus hard CE | page | -0.022 [-0.077; +0.033] | -0.033 [-0.058; -0.010]* | -0.111 [-0.158; -0.035]* |
| MSE without all-zero targets minus hard CE | page | -0.015 [-0.071; +0.040] | -0.040 [-0.065; -0.017]* | -0.099 [-0.145; -0.038]* |
| label smoothing minus hard CE | page | -0.021 [-0.054; +0.009] | +0.011 [-0.001; +0.023] | -0.071 [-0.109; -0.035]* |
| MSE minus label smoothing | page | -0.001 [-0.057; +0.051] | -0.044 [-0.065; -0.025]* | -0.039 [-0.077; +0.028] |
| KL minus soft CE | page | +0.013 [-0.011; +0.038] | -0.002 [-0.009; +0.005] | -0.012 [-0.033; +0.031] |
| soft CE minus hard CE | manuscript | -0.012 [-0.082; +0.053] | -0.028 [-0.049; -0.007]* | -0.055 [-0.115; +0.021] |
| KL minus hard CE | manuscript | +0.035 [-0.029; +0.099] | -0.030 [-0.052; -0.009]* | -0.031 [-0.103; +0.049] |
| MSE minus hard CE | manuscript | +0.000 [-0.064; +0.064] | -0.023 [-0.044; -0.003]* | -0.071 [-0.120; +0.023] |
| MSE without all-zero targets minus hard CE | manuscript | +0.041 [-0.023; +0.105] | -0.028 [-0.049; -0.008]* | -0.066 [-0.130; +0.025] |
| label smoothing minus hard CE | manuscript | +0.000 [-0.035; +0.035] | +0.017 [+0.006; +0.029]* | -0.047 [-0.106; +0.030] |
| MSE minus label smoothing | manuscript | +0.000 [-0.064; +0.058] | -0.040 [-0.059; -0.023]* | -0.024 [-0.073; +0.053] |
| KL minus soft CE | manuscript | +0.047 [+0.012; +0.088]* | -0.002 [-0.009; +0.005] | +0.024 [-0.055; +0.089] |

## A2. Sensitivity to the threshold that defines transitional pages.

Table A2. Sensitivity to the threshold that defines transitional pages. Differences in JS from hard cross-entropy on the pages whose maximum of the mapped distribution is below the threshold; in parentheses, p from the exact manuscript-level permutation test.

| threshold | pages | manuscripts | soft CE | KL | MSE |
|---|---|---|---|---|---|
| 0.50 | 57 | 7 | -0.138 (0.031) | -0.150 (0.016) | -0.135 (0.016) |
| 0.55 | 73 | 11 | -0.115 (0.031) | -0.122 (0.012) | -0.105 (0.046) |
| 0.60 | 97 | 16 | -0.111 (0.010) | -0.112 (0.005) | -0.115 (0.006) |
| 0.65 | 214 | 41 | -0.046 (0.056) | -0.050 (0.028) | -0.064 (0.003) |

The conclusion depends on the threshold quantitatively but not in sign. The stricter the threshold, the larger the gain in divergence, but the fewer manuscripts remain in the subsample: seven at a threshold of 0.5, eleven at 0.55 and sixteen at 0.6. At these three thresholds every difference is significant (p is at most 0.046). At a threshold of 0.65 (41 manuscripts) the difference for soft cross-entropy is no longer significant (p = 0.056), while the differences for KL and MSE are. The point estimate of the loss in argmax accuracy grows together with the gain. At a threshold of 0.5 it reaches -36.8 pp for soft cross-entropy, but even there it rests on four manuscripts out of seven. With that many manuscripts the permutation test cannot return a p value below 0.125, so the significance of this loss cannot be established.

A3. Direct outputs: paired differences. They are discussed in Section 5.3.

Table A3. Paired differences: the direct outputs minus the collapsed 14-dimensional vector. All 714 pages: cluster bootstrap over manuscripts, 10,000 replicates, 95%, asterisks mark intervals that exclude zero. Transitional pages (97 pages, 16 manuscripts): the difference, and in parentheses p from the exact manuscript-level permutation test.

| comparison | JS region, all | JS mode, all | JS region, transitional (p) | JS mode, transitional (p) |
|---|---|---|---|---|
| direct outputs minus hard CE | -0.009 [-0.026; +0.006] | -0.015 [-0.022; -0.007]* | -0.066 (0.081) | -0.005 (0.55) |
| direct outputs minus soft CE | -0.001 [-0.013; +0.009] | -0.008 [-0.014; -0.002]* | +0.028 (0.079) | -0.006 (0.29) |
| direct outputs minus label smoothing | -0.025 [-0.039; -0.012]* | -0.010 [-0.015; -0.005]* | -0.041 (0.16) | +0.001 (0.93) |

A4. Patch binarization. The first run is reported in Table 1. For the second run of the binarized S2 model (random seed 1) the S1K minus S2 gap is +35.7 pp [+30.5; +41.0] at the patch level, +42.0 pp [+34.7; +49.0] at the page level and +36.8 pp [+29.8; +43.9] at the manuscript level. At the patch level the gap is smaller than with grayscale input (35.5 against 46.6 pp). Binarization lowers the patch-level accuracy of S1K from 0.896 to 0.795, but after averaging over the page it barely changes it (0.975 against 0.972).

On unseen manuscripts the point estimate of accuracy is higher with binarization, but the gain is not established. We trained the binarized S2 model twice, with random seeds 0 and 1. Page-level accuracy is 0.582 and 0.555 against 0.529 to 0.541 for the three grayscale runs. The paired difference against the grayscale model trained on the same hardware with the same seed is +5.0 pp [+0.3; +9.9] at the page level for seed 0 and +2.7 pp [-3.0; +8.4] for seed 1. Of the six page-level comparisons of the two binarized runs with the three grayscale ones, two are significant, both for the first binarized run (+5.0 and +5.3 pp; against the original grayscale run on the Tesla P100 it is +4.1 pp, not significant), while for the second run the difference ranges from +1.4 to +2.7 pp. The bootstrap does not account for run-to-run variation here. This comparison was not part of the rule specified in advance (Section 5.1). On the official blind test set the point estimates with binarization are almost the same: 0.341 at the patch level, 0.407 at the page level and 0.439 at the manuscript level, against 0.334, 0.390 and 0.439. The paired page-level difference is +1.7 pp [-11.0; +14.5], a wide interval. Our pipeline already falls short of the figures published by the authors of the dataset with grayscale input: their ResNet50 classifies 52% of blind patches correctly against our 0.334. Otsu's thresholding in our pipeline does not close that gap (0.341 against their 62% with binarized input). Which of the remaining differences between the pipelines account for it we did not test.

With binarized input the model remains overconfident on unseen manuscripts at the patch level (mean confidence 0.831 against 0.441 accuracy, ECE 0.390 against 0.115 under S1K). At the page level the overconfidence is markedly smaller, 0.609 against 0.582. ECE is 0.063 against 0.172 for the grayscale model trained on the same hardware with the same seed (paired difference -0.109 [-0.157; -0.021]). Against the two other grayscale runs it is -0.100 and -0.105, also significant. The second binarized run gives 0.613 against 0.555, and its reduction in ECE relative to each of the three grayscale runs is significant (from -0.074 to -0.083).

Page-level confidence comes from averaging over patches and is close to the share of patches that agree with the page prediction (0.609 and 0.638), so text-free patches affect it as well. If a page is averaged only over the patches where writing is almost always present (the largest dark connected component covers no more than 10% of the patch area), the binarized S2 model is more overconfident: confidence exceeds accuracy by 0.117 [+0.051; +0.183] instead of 0.028, but the reduction in overconfidence relative to the grayscale model still holds (-0.115 [-0.166; -0.064]), and so does the reduction in ECE (-0.112 [-0.158; -0.059]). With binarization the S1K model is underconfident at the page level (0.826 against 0.975), and its ECE, which does not distinguish overconfidence from underconfidence, is higher than that of S2. The difference between mean confidence and accuracy is larger for S2 than for S1K with binarization too (S1K minus S2 is -0.177 [-0.242; -0.112], and -0.218 [-0.289; -0.147] with grayscale input). Confidence is still useful for deciding which attributions to re-examine: in the more confident half of the pages 77% of the predictions are correct, in the less confident half 40%. By the end of training the binarized models fit the training patches less closely (final-epoch loss from 0.0444 to 0.0585 against 0.0017 to 0.0026 for the grayscale models), so we attribute the smaller page-level overconfidence to binarization together with this difference in fit. On the official blind test set the page-level overconfidence after binarization is also smaller, but it does not disappear: 0.547 against 0.407, versus 0.661 against 0.390 for the grayscale model.

A5. Text-free patches. The gap barely depends on text-free patches. The automatic criterion flags a patch when the largest dark connected component covers more than 30% of the patch area after Otsu's thresholding, and by visual inspection 52 of 60 such patches carry no writing. If the flagged patches are dropped when a page is averaged, the page-level gap is +42.8 pp [+35.6; +49.8], and +39.2 pp [+31.9; +46.3] with binarization. The criterion finds only part of the text-free patches, about 42% by visual inspection. Therefore, we also checked a stricter selection. It keeps only the patches where writing is almost always present (the largest dark connected component covers no more than 10% of the patch area, which removes about 80% of the text-free patches). With it the page-level gap is +42.7 pp [+35.4; +49.7], and +39.3 pp [+32.1; +46.5] with binarization.

A6. Repeated runs. We repeated hard-label training on a Tesla T4 with random seeds 0 and 1 (the original run with seed 0 was carried out on a Tesla P100), keeping the same folds and the same patch selection. Across the three runs accuracy is 0.529 to 0.541 at the page level and 0.585 to 0.602 at the manuscript level (a range of 1.3 and 1.8 pp). Page-level ECE and JS are 0.163 to 0.172 and 0.290 to 0.294. The paired differences between runs are not significant. In our runs the hard model varies less across runs than the soft target (1.8 against 5.3 pp at the manuscript level), but with this many runs (three and four) that difference is not established. With the repeated S2 runs the gap between protocols is 44.1 and 44.3 pp at the page level and 40.9 and 39.2 pp at the manuscript level.

We compared each of the four soft-target runs (cross-entropy with soft targets and KL) with each of the three runs of the hard model. JS for the soft target is significantly lower in all twelve pairs at both levels. Page-level ECE is lower in all twelve pairs and significantly so in eleven. At the manuscript level the difference is not significant in any pair. We found no evidence of an accuracy difference in any pair, but at the page level the point estimate for the soft target is lower in all twelve (from -4.2 to -0.7 pp). Averaged over runs it is 0.509 against 0.534, and all four soft-target runs lie below all three runs of the hard model. With seven runs such a complete separation gives the smallest p attainable in a permutation test over runs (2/35, about 0.06), so a small loss of page-level accuracy cannot be ruled out. For MSE the page-level gain in ECE is significant in every pair, twelve out of twelve for plain MSE and nine out of nine without all-zero targets. JS for plain MSE is significantly lower in eleven of the twelve pairs at the page level and in six at the manuscript level. Without all-zero targets it is lower in all nine pairs at both levels. Label smoothing is worse than hard-label training in manuscript-level JS in two pairs out of three. The qualitative conclusions from the comparison with hard-label training do not depend on which of its runs is chosen, and only the number of significant pairs changes.

A7. MSE without all-zero targets. The calibration gain of MSE might be explained by the pages with an all-zero vector, since for MSE their all-zero target pulls the output toward the uniform distribution (Section 3). We tested this by excluding all-zero targets from the loss. We repeated training with three random seeds, and for comparison we trained plain MSE on the same hardware with the same seeds. The decision rule was specified in advance, before the results were available. Without all-zero targets, ECE is 0.240 to 0.244 at the patch level, 0.057 to 0.069 at the page level and 0.076 to 0.097 at the manuscript level. The difference from hard-label training is significant in all three runs at the patch level (from -0.200 to -0.204) and at the page level (from -0.094 to -0.106), while at the manuscript level the intervals include zero. Across the four runs of plain MSE, ECE is 0.209 to 0.231 at the patch level and 0.052 to 0.062 at the page level. Table 3 reports the run with the lowest values at these levels. We found no evidence of a difference in accuracy or JS against plain MSE. The all-zero targets do lower confidence on their own 25 pages (0.542 to 0.596 against 0.611 to 0.651 without them), but the calibration gain of MSE survives their exclusion. In patch-level ECE all three runs without all-zero targets are significantly lower than every run of soft cross-entropy and of KL (differences from -0.052 to -0.068). At the page level they are lower by the point estimate, but the paired differences against them are not significant.

The variant without all-zero targets differs from plain MSE by +0.020 on average at the patch level and by +0.004 at the page level. A paired t-test on the three differences would rest on only two degrees of freedom, so the rule instead compares the mean difference with the run-to-run variation of a single run, estimated from the repeated runs of the soft target and of plain MSE, and it treats two runs that share a random seed as independent. That is an assumption. The two runs share the data order and the initialization, but because the loss differs their training trajectories diverge from the first step. For independent runs the standard deviation of the difference between two of them equals the variation of a single run multiplied by the square root of two. For the mean of three such differences it equals that variation multiplied by the square root of 2/3; the threshold is 2.6 times this last quantity, that is 0.022 at the patch level and 0.028 at the page level. The multiplier 2.6 is close to the critical value of t for a two-sided 5% level with five degrees of freedom: that is how many the pooled estimate of the variation has over the four runs of the soft target and the three runs of plain MSE that the rule assumed when it was written down. The variation of plain MSE that enters the threshold is itself computed from all four of its runs. The rule takes not the pooled estimate but the larger of the two, so the threshold is higher than it would be with the pooled estimate, which makes a difference harder to detect. Strictly, such a threshold applies only if the run-to-run variation is the same for all training variants. Three or four runs are not enough to check that, so we also computed other ways of estimating the variation. At the page level the difference is far from the threshold under every one of them. At the patch level the conclusion is borderline: the 5% thresholds for the mean difference range from 0.017 to 0.027 across these methods, and the paired t-test on the three differences themselves gives 0.022, the same as the rule. The observed +0.020 falls inside that range, and p ranges from 0.025 to 0.095 (0.057 for the paired t-test). In all three pairs the ECE without all-zero targets is slightly higher, and in two of them the paired interval excludes zero. Therefore, we cannot rule out that calibration at the patch level is slightly worse without all-zero targets.

## REFERENCES

Reference formatting will follow the requirements of the target venue.

Data: VML-HP-ext dataset, version 2, Zenodo, doi:10.5281/zenodo.6856934, licensed CC-BY-4.0.